\documentclass[11pt]{article}
\usepackage[margin=1in]{geometry}
\usepackage{hyperref}
\usepackage{booktabs}

\title{Part Grounding, Not Action Knowledge:\\
Locating the Bottleneck in VLM Affordance Prediction}
\author{Sarthak Sattigeri\\
\small Manipal University Jaipur\\
\small \url{https://github.com/S1rlvk/vlm-physical-affordances}}
\date{}

\begin{document}
\maketitle

\begin{abstract}
Benchmarks agree that vision-language models reason poorly about low-level manipulation, but an aggregate accuracy does not say which step fails. We separate two steps that affordance questions usually conflate: identifying which part of an object is the one to act on, and knowing what action that part requires. On 19 articulated objects we asked eight models, spanning three developers, what motion a robot should apply. Under an open prompt, \texttt{push} was produced once in 64 evaluations where it was correct, seven of the eight models never producing it at all, despite \texttt{push} being correct for 8 of the 19 objects and appearing in the offered label set every time. That looks like a hole in the models' action vocabulary. It is not. Inspecting the outputs showed the models were describing a different part than the one being scored: asked what to do with a camera, they explained how to pick up the camera, and we marked them wrong for not discussing its button. Naming the target part in the prompt raises action accuracy by 0.32 to 0.63 for every model, from a range of 0.158--0.474 to 0.684--0.947, and \texttt{push} recall from 0--1/8 to 7--8/8. Under the open prompt no model beats a constant answer that ignores the image; once the part is named, all eight do. Asked to describe the same part in free prose with no label set, models produce pressing language for 6 to 8 of the 8. These results are hard to reconcile with an account on which the failures reflect missing action knowledge, and point instead to part grounding as the dominant bottleneck in this evaluation. The pattern holds across all three families and does not diminish with model capability. We are careful about what condition C establishes: naming the part supplies the grounding variable, so it bounds what a perfect part detector would offer rather than demonstrating a general model of mechanics. Two supporting results point the same way: on real photographs only three of eight models localize grasp points better than a constant that ignores the image, and on rendered objects none do. We also document two measurement errors of our own, a threshold that let a constant baseline score 0.929 and a labelling rule wrong on 4 of 19 objects, both caught only by testing our numbers against trivial alternatives.
\end{abstract}

\section{Introduction}

A vision-language model that names a drawer has not told a robot anything it can act on. Manipulation needs to know where to make contact and which way the part moves.

Existing benchmarks establish that models do this badly. ManipBench \cite{manipbench} finds articulated-object manipulation its hardest category at 0.396 mean accuracy across 33 model variants, against roughly 0.99 for humans. Work on affordance inference for non-humanoid robots reports models omitting a correct affordance in favour of a more common action from training \cite{nonhumanoid}.

We ask which step is failing. An affordance question bundles two inferences: find the part that matters, then say what that part does. A single accuracy figure cannot separate them, and they have different consequences. If models lack physical knowledge, a robotics pipeline cannot rely on them for action selection. If they hold the knowledge but attend to the wrong part, a part detector placed in front of the VLM fixes the problem, and the VLM is useful after all.

Our result is that the second explanation is the right one. The effect is large, it appears in every model we tested from three independent developers, and it does not shrink as models get more capable.

\section{Method}

We evaluate on 19 articulated objects from GAPartNet \cite{gapartnet}, a re-annotation of PartNet-Mobility \cite{partnetmobility}, rendered with a fixed camera in their default closed pose, plus 28 real tool photographs from the UMD Part Affordance dataset \cite{umd} used for the spatial task.

Action labels map deterministically from GAPartNet's annotated part category: \texttt{slider\_button} to \texttt{push}, \texttt{slider\_drawer} and \texttt{hinge\_door} to \texttt{pull}, \texttt{slider\_lid} to \texttt{lift-vertical}. Parts whose action is ambiguous from one static view, such as knobs, map to no label and are excluded rather than guessed. The resulting distribution is \texttt{pull} 10, \texttt{push} 8, \texttt{lift-vertical} 1. We release the full audit table.

We evaluate eight models from three developers: Qwen3-VL-8B (open-weight, run locally); gpt-4o-mini, gpt-5-mini and gpt-5; and Claude Haiku 4.5, Sonnet 5, Opus 5 and Fable 5. Each is run under three prompts, holding the image fixed:

\begin{itemize}
\item[\textbf{A}] The model chooses which part to discuss; the answer is constrained to the five labels. This is the standard framing.
\item[\textbf{C}] We name the target part; the answer is constrained to the five labels.
\item[\textbf{B}] We name the target part; the model answers in free prose with no labels offered.
\end{itemize}

A against C isolates part selection from action knowledge. C against B tests whether the fixed label set suppresses an answer the model would otherwise give. Scoring for B uses keyword sets fixed before the run and written to be generous toward \texttt{push}, which biases against the hypothesis we held going in.

\section{Results}

\subsection{An apparent blind spot}

Under prompt A, \texttt{push} appeared once in 64 model-object evaluations where it was the correct answer (Table \ref{tab:ablation}), a single prediction from gpt-5; the other seven models never produced it. This is on a task where \texttt{push} was correct for 8 of 19 objects and appeared in the offered label set on every call. Accuracy ran from 0.158 to 0.474, and no model reached the 0.526 that a constant \texttt{pull} answer achieves without looking at the image.

Read on its own this suggests the models cannot represent pushing. Inspecting the outputs says otherwise. On all 8 push objects, every model named a part other than the one being scored: "camera body", "paper sheet", "top edge of body", "lid front edge". Never a button. The models were answering how to pick the object up, a reasonable reading of a question about what a robot should do with a camera, and our scoring counted it wrong.

\subsection{Naming the part removes it}

\begin{table}[h]
\centering
\begin{tabular}{llccc|ccc}
\toprule
& & \multicolumn{3}{c|}{Action accuracy} & \multicolumn{3}{c}{\texttt{push} recall (of 8)} \\
Model & Family & A & C & $\Delta$ & A & C & B \\
\midrule
Qwen3-VL-8B      & Alibaba   & 0.158 & 0.737 & $+0.58$ & 0 & 8 & 7 \\
gpt-4o-mini      & OpenAI    & 0.263 & 0.842 & $+0.58$ & 0 & 8 & 8 \\
gpt-5-mini       & OpenAI    & 0.316 & 0.684 & $+0.37$ & 0 & 8 & 8 \\
gpt-5            & OpenAI    & 0.474 & 0.947 & $+0.47$ & 1 & 8 & 8 \\
Claude Haiku 4.5 & Anthropic & 0.474 & 0.789 & $+0.32$ & 0 & 7 & 6 \\
Claude Sonnet 5  & Anthropic & 0.158 & 0.789 & $+0.63$ & 0 & 8 & 7 \\
Claude Opus 5    & Anthropic & 0.368 & 0.789 & $+0.42$ & 0 & 8 & 8 \\
Claude Fable 5   & Anthropic & 0.368 & 0.895 & $+0.53$ & 0 & 8 & 8 \\
\midrule
Constant \texttt{pull} & --- & 0.526 & 0.526 & --- & 0 & 0 & --- \\
\bottomrule
\end{tabular}
\caption{Prompt A lets the model choose the part; C names it; B names it and asks for prose. 19 objects per cell, 456 calls total, no failures. Under A no model reaches the constant baseline; under C all eight exceed it.}
\label{tab:ablation}
\end{table}

Naming the part raises accuracy by 0.32 to 0.63 for every model, and recovers \texttt{push} almost completely: seven of eight reach 8/8 and Haiku 4.5 reaches 7/8. All eight exceed the constant baseline under C, which none did under A. The effect appears in all three families and does not diminish with capability. gpt-5 is the strongest model in the study under C at 0.947, and still sits at 0.474 when it must choose the part itself.

The label set is not the constraint either. Under B, with no options offered and prose requested, models produce pressing language for 6 to 8 of the 8 push objects, including 7/8 for the open-weight model, describing buttons that move "inward (depresses) when pressed and springs back". The vocabulary was always available.

The pattern does not track model capability in the way a knowledge deficit would. Under A the eight models spread across 0.32 accuracy points and none clears the baseline; under C all eight clear it. The improvement is roughly uniform across families and sizes, from an 8B open-weight model to the largest proprietary ones, which is what one expects if the missing ingredient is knowing which part to describe rather than knowing what parts do.

\subsection{Spatial results point the same way}

If part grounding is the weak step, spatial localization should be weak too. Table \ref{tab:spatial} reports it two ways for all eight models, and the pair is instructive.

\begin{table}[h]
\centering
\begin{tabular}{lcccc}
\toprule
& \multicolumn{2}{c}{Photographs ($n=28$)} & \multicolumn{2}{c}{Renders ($n=19$)} \\
Model & Hit rate & Mean dist. & Hit rate & Mean dist. \\
\midrule
Qwen3-VL-8B      & 1.000 & \textbf{0.0356} & 0.684 & 0.1744 \\
gpt-4o-mini      & 0.786 & 0.1367 & 0.737 & 0.2069 \\
gpt-5-mini       & 0.929 & 0.1327 & 0.368 & 0.2693 \\
gpt-5            & 1.000 & \textbf{0.0631} & 0.421 & 0.2363 \\
Claude Haiku 4.5 & 0.750 & 0.1620 & 0.368 & 0.2485 \\
Claude Sonnet 5  & 0.429 & 0.1765 & 0.579 & 0.2089 \\
Claude Opus 5    & 1.000 & 0.0955 & 0.421 & 0.2262 \\
Claude Fable 5   & 1.000 & \textbf{0.0226} & 0.474 & 0.2226 \\
\midrule
Constant $(0.5,0.5)$ & 0.929 & 0.0968 & 0.947 & 0.1438 \\
\bottomrule
\end{tabular}
\caption{Spatial localization, all 47 objects per model. Hit rate uses a 0.15 threshold on the image diagonal; mean distance is normalized Euclidean error, lower is better. The constant baseline ignores the image. Bold marks the three models that clearly beat it on photographs.}
\label{tab:spatial}
\end{table}

By hit rate, four models score perfectly on photographs and the task looks solved. By mean distance, three of the eight clearly beat a constant that never looks at the image: Fable 5 (0.0226), Qwen3-VL-8B (0.0356) and gpt-5 (0.0631). Opus 5 is the instructive case, pairing a hit rate of 1.000 with a mean distance of 0.0955 against the baseline's 0.0968, so its perfect score and the constant's near-perfect score measure the same thing. The remaining four models are worse than answering the center every time. On renders no model beats the baseline on either measure; the best, Qwen3-VL-8B at 0.1744, still trails the constant's 0.1438.

The rendered column is the one that matters for our argument, because those are the images the action task uses. There, localization is poor across the board, which is consistent with part grounding being the step that fails.

\subsection{Two errors in our own measurements}

We report these because both produced credible numbers that were wrong, and both were caught only by comparison against something trivial.

The spatial threshold was set by hand at 0.15 of the image diagonal before we had any model output. Because these datasets center their objects, ground-truth points span 0.32 in $x$ and 0.14 in $y$, narrower than the threshold itself. A model answering $(0.5,0.5)$ every time scores 0.929 on photographs and 0.947 on renders. Our first spatial result, three of five models scoring a perfect hit rate, was largely a measurement of object centering.

The action labels were originally derived from joint geometry, using a hand-chosen travel threshold of 0.02 units to separate buttons from drawers and an alignment test against $(0,0,1)$. That rule was wrong on 4 of 19 objects. Every \texttt{slider\_button} fell under the travel threshold except a keyboard key at 0.025, which became \texttt{pull}. Two \texttt{slider\_drawer} parts became \texttt{lift-vertical} because their axes aligned with $(0,0,1)$ in object-local coordinates, a frame whose relation to world-up we never checked. Correcting these raised every model's accuracy, so the errors had been penalizing the models rather than supporting our hypothesis. The fix was to stop inferring from geometry what the dataset annotation already stated.

\section{Discussion}

The practical reading is that these models are more useful for manipulation than an open-prompt evaluation suggests, provided something else selects the part. Accuracy near 0.8 on a five-way action classification is a usable signal. Accuracy near 0.3, which is what the same models produce when asked to find the part themselves, is not. A pipeline that runs part segmentation first and queries the VLM about a named part is doing substantially different work from one that hands the VLM an image and a goal.

It also suggests that affordance evaluations may be measuring part grounding more than action selection without distinguishing the two. Our own result looked like a clean finding about a missing action category until we varied the prompt, and the correction moved the effect from one class of explanation to another entirely. We describe this as affordance prediction and part grounding rather than physical reasoning, because our labels map part categories onto conventional actions and a model can satisfy them without any general account of mechanics.

The same caution applies to our earlier work on grasp-force prediction, where a model produced force estimates spanning 6 distinct values against 44 in the ground truth and we read this as absent physics. A rival explanation we did not test is that the model was answering a coarser question than the one being scored.

\section{Limitations}

The ablation and the spatial analysis both cover eight models from three developers, and the part-naming effect appears in all of them, which makes shared training data an unlikely explanation.

Nineteen articulated objects supports no per-model claims and we make none; the effect we report is large and consistent in direction rather than precisely estimated. Naming a part supplies information, so condition C is an upper bound on what a perfect part detector would provide rather than a measurement of one. Our labels are a deterministic map from existing part annotations rather than fresh human annotation, so the task reduces to identifying the part type and applying its conventional action; under prompt A the models did not achieve that. Accuracies under prompt A also differ slightly from an earlier prompt variant that requested additional fields, so these numbers carry prompt sensitivity of their own.

\section{Conclusion}

Across eight vision-language models from three developers, one of five offered action labels was produced once in 64 relevant evaluations when the models chose which part to describe, and for 7 or 8 of 8 once they were told. Action accuracy rose by 0.32 to 0.63 for every model under the same change, and every model moved from below a constant baseline to above it. We read this as evidence against an interpretation on which these failures are primarily missing action knowledge, and as pointing to part grounding as the step worth fixing. We stop short of claiming the models hold a general model of mechanics: condition C hands them the grounding variable, and our labels reduce to mapping a part category onto its conventional action.

\appendix
\section{Action label audit}

Every action label is a deterministic function of the part category in column three, via the mapping in Section 2. Joint type, limits and axis are listed so the mapping can be checked against the underlying mechanics independently. Auditing this table is what surfaced the four errors described in Section 3.4. The axis column shows why an earlier rule that tested alignment against $(0,0,1)$ failed: those axes are expressed in each object's own frame, not a shared world frame. The machine-readable version ships with the code as \texttt{data/action\_label\_audit.json}.

\begin{table}[h]
\centering
\small
\begin{tabular}{lllllll}
\toprule
Object & ID & Part category & Joint & Limits & Axis & Label \\
\midrule
Safe & 101613 & \texttt{hinge\_door} & hinge & 0, -159.12 & (0.00, -1.00, 0.00) & \texttt{pull} \\
Camera & 102505 & \texttt{slider\_button} & slider & 0, 0.008 & (-0.99, 0.13, 0.00) & \texttt{push} \\
Remote & 101015 & \texttt{slider\_button} & slider & 0, 0.01 & (0.00, 0.00, -1.00) & \texttt{push} \\
Refrigerator & 11178 & \texttt{hinge\_door} & hinge & 0, -180 & (0.00, -1.00, 0.00) & \texttt{pull} \\
Printer & 104016 & \texttt{slider\_button} & slider & 0, 0.004 & (-0.00, -0.99, -0.15) & \texttt{push} \\
WashingMachine & 103425 & \texttt{hinge\_door} & hinge & 0, -97.92 & (0.00, 1.00, 0.00) & \texttt{pull} \\
Keyboard & 13082 & \texttt{slider\_button} & slider & 0, -0.025 & (0.00, 1.00, 0.00) & \texttt{push} \\
Phone & 103347 & \texttt{slider\_button} & slider & 0, 0.005 & (0.00, 0.00, -1.00) & \texttt{push} \\
Oven & 101943 & \texttt{hinge\_door} & hinge & 0, -90 & (-1.00, 0.00, 0.00) & \texttt{pull} \\
Dishwasher & 12560 & \texttt{hinge\_door} & hinge & 0, 90 & (1.00, 0.00, 0.00) & \texttt{pull} \\
Table & 22301 & \texttt{slider\_drawer} & slider & 0, 0.596 & (0.00, 0.00, 1.00) & \texttt{pull} \\
KitchenPot & 100021 & \texttt{slider\_lid} & slider & 0, 0.05 & (0.00, 1.00, 0.00) & \texttt{lift-vertical} \\
Toilet & 102682 & \texttt{slider\_button} & slider & 0, -0.01 & (0.00, 1.00, 0.00) & \texttt{push} \\
StorageFurniture & 48740 & \texttt{slider\_drawer} & slider & 0, 0.6 & (0.00, 0.00, 1.00) & \texttt{pull} \\
Safe & 101599 & \texttt{hinge\_door} & hinge & 0, 159 & (0.00, 1.00, 0.00) & \texttt{pull} \\
Camera & 102845 & \texttt{slider\_button} & slider & 0, -0.004 & (0.00, 1.00, 0.00) & \texttt{push} \\
Remote & 100270 & \texttt{slider\_button} & slider & 0, 0.004 & (0.00, 0.00, -1.00) & \texttt{push} \\
Door & 9288 & \texttt{hinge\_door} & hinge & 0, 90 & (0.00, 1.00, 0.00) & \texttt{pull} \\
Refrigerator & 12054 & \texttt{hinge\_door} & hinge & 0, -180 & (0.00, -1.00, 0.00) & \texttt{pull} \\
\bottomrule
\end{tabular}
\caption{All 19 articulated objects with the mechanics their labels derive from.}
\end{table}


\begin{thebibliography}{9}
\bibitem{manipbench} ManipBench: Benchmarking Vision-Language Models for Low-Level Robot Manipulation. CoRL, 2025. arXiv:2505.09698.
\bibitem{nonhumanoid} Assessing VLM-Driven Semantic-Affordance Inference for Non-Humanoid Robot Morphologies. arXiv:2604.19509.
\bibitem{umd} A. Myers, C. L. Teo, C. Fermuller, Y. Aloimonos. Affordance Detection of Tool Parts from Geometric Features. ICRA, 2015.
\bibitem{gapartnet} H. Geng et al. GAPartNet: Cross-Category Domain-Generalizable Object Perception and Manipulation via Generalizable and Actionable Parts. CVPR, 2023.
\bibitem{partnetmobility} F. Xiang et al. SAPIEN: A SimulAted Part-based Interactive ENvironment. CVPR, 2020.
\end{thebibliography}
\end{document}